\pdfoutput=1
\documentclass[11pt]{article}
\usepackage[table]{xcolor} %new added

\usepackage[final]{acl}

\usepackage{times}
\usepackage{latexsym}

\usepackage[T1]{fontenc}
\usepackage[utf8]{inputenc}

\usepackage{microtype}

\usepackage{inconsolata}

\usepackage{graphicx}

\usepackage{array,multirow}
\usepackage{arydshln} % dashed line in table
\usepackage{amssymb}
\usepackage{pifont} % to insert emoji, not the real one
\usepackage{float}
\usepackage{amsfonts} %load this before mathabx
\usepackage{amsmath}

\usepackage{booktabs} % \toprule shows in table
\usepackage{hyperref}
\usepackage{cleveref} % clever refm for chapter symbole
\crefformat{section}{\S#2#1#3} % see manual of cleveref, section 8.2.1
\crefformat{subsection}{\S#2#1#3}
\crefformat{subsubsection}{\S#2#1#3}
\usepackage{url}
\usepackage{tikz-dependency}
\usepackage{mathabx}
\usepackage{tabularray}
\usepackage{tcolorbox}
\usepackage{lipsum}
\usepackage{etoolbox}
\AtBeginEnvironment{tcolorbox}{\small}
\makeatletter
\newcommand*\bigcdot{\mathpalette\bigcdot@{.5}}
\newcommand*\bigcdot@[2]{\mathbin{\vcenter{\hbox{\scalebox{#2}{$\m@th#1\bullet$}}}}}
\makeatother

\title{Topic-Guided Reinforcement Learning with LLMs for Enhancing Multi-Document Summarization}

\author{
Chuyuan Li$^{1}$, Austin Xu$^{2}$, Shafiq Joty$^{2}$, Giuseppe Carenini$^{1}$ \\ 
$^{1}$ Department of Computer Science, University of British Columbia \\
$^{2}$ Salesforce AI Research \\ 
\texttt{chuyuan.li@ubc.ca}, \ \texttt{carenini@cs.ubc.ca} \\ 
\texttt{\{austin.xu, sjoty\}@salesforce.com}
}

\begin{document}
\maketitle
\begin{abstract}

A key challenge in Multi-Document Summarization (MDS) is effectively integrating information from multiple sources while maintaining coherence and topical relevance.
While Large Language Models have shown impressive results in single-document summarization, their performance on MDS still leaves room for improvement. 
In this paper, we propose a topic-guided reinforcement learning approach to improve content selection in MDS.  
We first show that explicitly prompting models with topic labels enhances the informativeness %and factual accuracy 
of the generated summaries. 
Building on this insight, we propose a novel topic reward within the Group Relative Policy Optimization (GRPO) framework to measure topic alignment between the generated summary and source documents.
Experimental results on the Multi-News and Multi-XScience 
datasets demonstrate that our method consistently outperforms strong baselines, highlighting the effectiveness of leveraging topical cues in MDS.
\footnote{Our code is available at \url{https://github.com/chuyuanli/TopicRL-for-MDS}.}
\end{abstract}

%================
\section{Introduction}

\begin{figure}[t]
    \centering
    \includegraphics[width=\columnwidth]{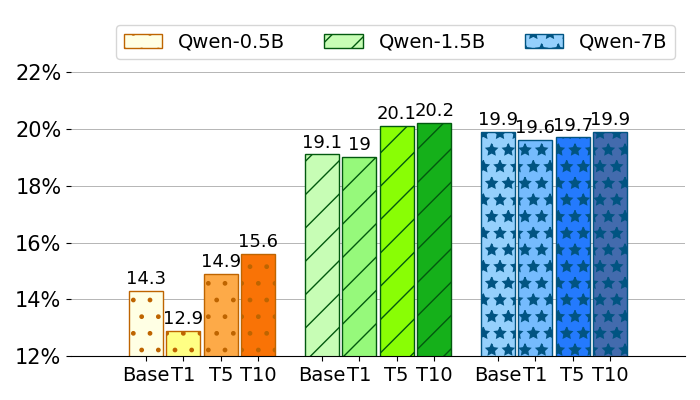}
    \caption{Performance on Multi-News \citep{fabbri2019multi} using prompting (Base) and topic-incorporated prompting (T$n$; $n$ means number of topic labels) with Qwen2.5-series model \citep{qwen2025qwen25technicalreport}. The geometric mean of Rouge-1/2/L scores are reported. Topic key words are previously generated using a \textit{teacher} model: Qwen2.5-7B. 
    We see that topic-enhanced instruction (T$5$ and T$10$) improves small LLMs' (0.5B and 1.5B) performances over standard prompt (Base).}
    \label{fig:prompt-hs}
    \vspace{-2ex}
\end{figure}

\begin{figure*}[t]
    \centering
    \includegraphics[width=0.9\linewidth]{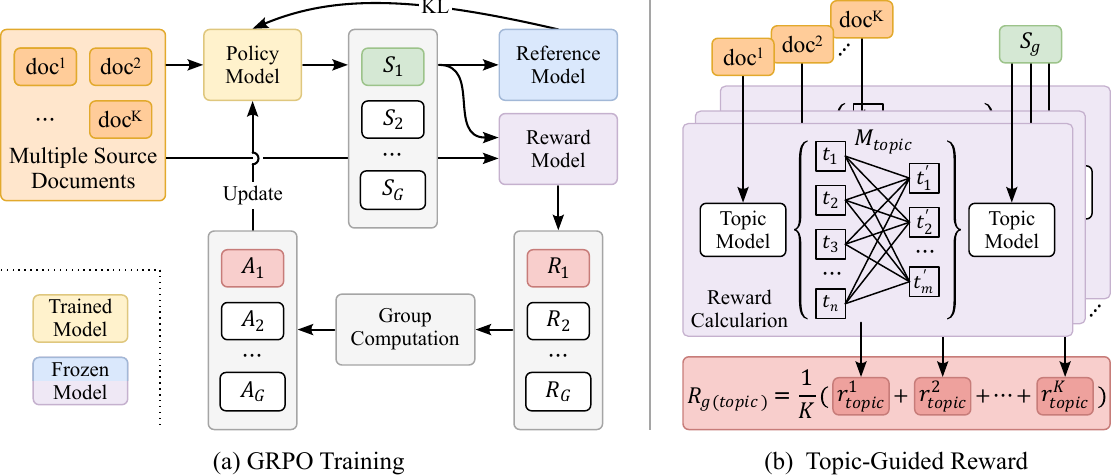}
    \caption{Multi-Document Summarization training (a) using our proposed Topic-Guided reward (b), with GRPO \citep{shao2024deepseekmath}.
    %For every input data that contains $K$ source documents
    %($\text{doc}^1$, \dots, $\text{doc}^K$),
    Every input data contains $K$ source documents
    ($\text{doc}^1$, \dots, $\text{doc}^K$).
    %we use a topic model to extract $I$ key topic phrases $\{t_1, t_2,\dots,t_I\}$ from each of them.
    For each document $\text{doc}^k$, we use a topic model to extract $n$ number of key topic phrases $\{t_1, t_2,\dots,t_n\}$.
    %For each completion (e.g., the generated summary $S_g$), 
    % \shafiq{use subscript $k$?}), 
    % To calculate the topic-guided reward of each completion,
    % (e.g., the generated summary $S_g$), 
    %
    Similarly,
    we extract $m$ topic phrases $\{t'_1, t'_2,\dots,t'_m\}$ from each generated summary $S_g$.
    %we also extract $J$ topic phrases $\{t'_1, t'_2,\dots,t'_J\}$. 
    %We construct a similarity matrix $M_g^k$ between a document-completion pair ($\text{doc}^k$-$S_g$), from which we compute a topic alignment score $r^k_{topic}$.
    %
    % We construct a topic similarity matrix $M_g^k$ 
    % from the $n$ and $m$ topic phrases of each document-summary pair.
    We construct a topic similarity matrix $M_g^k$ by comparing the $n$ and $m$ topic phrases from each source document-summary pair,
    % (i.e., total $K$ matrices for an input data).
    % We then compute a topic alignment score $r^k_{g\,\mathrm{(topic)}}$ using $M_g^k$.
    % Using this matrix, we compute a topic alignment score $r^k_{g\,\mathrm{(topic)}}$.
    from which we compute a topic alignment score $r^k_{g\,\mathrm{(topic)}}$.
    %The average of all $K$ topic scores in that input data ($R_{g(\text{topic})}$)
    % combined with a length-penalty reward (not shown in the figure), 
    We average the alignment scores over all $K$ document-summary pairs to derive the overall topic-guided reward $R_{g\,\mathrm{(topic)}}$, which is then used to calculate the group advantage $A_g$ for updating the policy model.
    % topic alignment scores to obtain the topic-guided reward $R_{g\,\mathrm{(topic)}}$ and 
    % use it
    % to compute the group advantage $A_g$ for updating the policy model.
    % \austin{Typo in (a) ``GRPO Traning''}--> fixed
    }
    \label{fig:pipeline}
    \vspace{-2ex}
\end{figure*}

Multi-Document Summarization (MDS) aims to generate a concise and coherent summary that captures the salient information from a collection of related documents. While recent advances in Large Language Models (LLMs) and prompting strategies have significantly improved the performance of abstractive summarization systems, existing MDS methods still struggle to maintain content relevance, coherence \citep{belem2024single}, and topic consistency \citep{amar2023openasp}, especially when synthesizing information across multiple sources \citep{liu2024benchmarking, lior2024seam}.

One important yet relatively underexplored direction in MDS is the incorporation of high-level %topic 
discourse information to guide the summarization process. 
Topics offer a global %semantic \shafiq{I'd say discourse} 
discourse structure that can help models identify salient content, resolve ambiguity, and enhance coherence in the generated summaries \citep{haghighi2009exploring, ouyang2007developing, li2023facing}. 
Early work incorporated topic distributions as auxiliary features to enrich word and sentence representations, either via topic models or graph-based approaches \citep{wei2012document, narayan2018don, wang2020friendly}. 
However, these methods typically operate at the token or sentence level and do not fully leverage topic signals as explicit guidance. 
More recent efforts have attempted to better align topic modeling with the summarization objective—for example, by using latent topics to pre-select salient sentences in extractive settings \citep{cui2020enhancing}, or by jointly learning topic representations and summarization 
% in a multi-task learning framework 
\citep{cui2021topic}. 
% \shafiq{Should we also mention this line of work:\url{ https://direct.mit.edu/tacl/article/doi/10.1162/tacl_a_00438/108867/Planning-with-Learned-Entity-Prompts-for} and  \url{https://aclanthology.org/2023.tacl-1.55/}; they did it for single-doc summ.}
Related work has also explored creating intermediate plans to guide summarization, such as creating \textit{Entity Chains} as key phrases \citep{narayan2021planning} or building question-answering blueprints \citep{narayan2023conditional}.
Despite these advances, the explicit use of topic labels as prompts or rewards to guide multi-document summarization remains largely unexplored.

%liu2019hierarchical paragraph-representation

% To bridge the gap between topic-guided summarization 
In this work, we investigate the role of explicit topic guidance in enhancing generic MDS. Different from previous studies that incorporate topic distributions or learns latent topics via neural topic modeling, we propose a more direct and interpretable strategy: guiding summarization models using topic phrases %that are 
explicitly extracted from the source documents. 
We begin with a simple yet insightful observation: prompting LLMs with extra topic information improves MDS quality in terms of informativeness. 
Figure~\ref{fig:prompt-hs} shows that 
when small LLMs (Qwen2.5-0.5B and 1.5B) are applied to summarization tasks, they show notable improvements if
prompted with topic labels (``T$5$'' and ``T$10$''), compared to using standard summarization prompt (``Base''). 
This motivates us to go beyond static prompting and incorporate topic awareness more directly into the training objective.

To this end, we introduce a novel reference-free topic-reward function that quantifies how well a generated summary aligns with its intended topics derived from each source document, see Figure~\ref{fig:pipeline} for an overview.
Our key assumption is that increasing the topical similarity between the generated summary and source documents will in turn improve the quality of summary generations. 
Accordingly, our reward is defined with respect to the improvements from (1) coverage: how well the generated summary covers important topics in source documents, and (2) precision: how relevant the topics in summary are to the source documents.
The final reward signal is a harmonic mean, %of topic coverage and precision,
which is then integrated into the Group Relative Policy Optimization framework \citep{shao2024deepseekmath} to enable reinforcement learning with topic-guided feedback. 

Specifically, we employ Qwen2.5-7B \citep{qwen2025qwen25technicalreport} within
the reward model to generate topic labels for a given document--either a source article or a summary--while using the smaller Qwen2.5-0.5B model as the policy model.
This setup also mirrors a knowledge distillation paradigm where the larger language model transfers topic-related knowledge to the smaller model to guide its learning process. 
We evaluate our method on two widely-used %MDS 
datasets: Multi-News \citep{fabbri2019multi} and Multi-XScience \citep{lu2020multi}, and demonstrate that our topic-aware training strategy leads to consistent improvements over standard and Reinforcement Learning from Human Feedback (RLHF)-guided baselines, as measured by both informativeness metrics (e.g., \textsc{Rouge}, \textsc{Llm} score) and topic alignment evaluation.

In summary, %our contributions are threefold:
(1) we show that using topic information improves MDS performance, both via prompting and LLM-based RL;
(2) we introduce a novel topic reward to measure source-summary discourse alignment, %This reward 
which is integrated into GRPO to perform topic-guided summarization;
% with reinforcement learning;
(3) empirical results indicate that topic-level signals represent a valuable yet underexploited form of supervision, yielding even stronger performance when combined with reference-based rewards like \textsc{Rouge}.

%================
\section{Related Work}

\paragraph{Multi-Document Summarization (MDS)}
Earlier approaches for MDS relies on extractive methods that rank and select salient sentences across documents \cite{nenkova2005impact, erkan2004lexrank, zhao2022read}. 
More recent work has shifted toward neural abstractive models that can generate coherent and fluent summaries from scratch \citep{liu2019hierarchical, zhang2020pegasus, ma2022multi}, while facing challenges in maintaining factual accuracy and topical consistency.
Alternative methods leverage task-specific pre-training like PRIMERA \citep{xiao2022primera} and graph networks \citep{liu2024glimmer, wang2022multi}. These approaches each target specific challenges (such as improve scientific MDS \citep{wang2022multi}) that are mostly orthogonal but complementary to our contributions on the general-purpose LLMs.

\paragraph{Discourse-Guided Summarization}
Although topic modeling has been widely used for document-level content understanding, its application to summarization has been relatively limited \citep{cui2021topic}.
\citet{harabagiu2005topic} explored topic representations using semantic themes (e.g., predicate-argument structures), which required a semantic parser to extract this structural information;
\citet{haghighi2009exploring} used LDA-style probabilistic topic models \citep{blei2003latent} to select topic-relevant sentences and showed improvement in terms of redundancy; \citet{cohan2018discourse} and \citet{wang2020friendly} introduced discourse-level and topic-aware attention mechanisms %in neural models 
to enhance long document summarization.
Another line of work involves discourse-level planning, where models generate summaries conditioned on given keywords \citep{he2022ctrlsum, dou2021gsum}, entities \citep{narayan2021planning}, or high-level concept \citep{zhong2021qmsum}. 
These approaches aim to control the focus of the summary based on %specific 
user intent or query,
while our work focuses on %generating 
generic summaries that holistically represent the source content using topical information.

\paragraph{Reinforcement learning (RL) for Summarization}

RL methods has been applied to summarization more broadly to optimize non-differentiable objectives, such as \textsc{Rouge} \citep{ranzato2016sequence, paulus2018deep, narayan2018don, wang2018reinforced} or human preferences \citep{ziegler2019fine, stiennon2020learning, ouyang2022training}.
Other approaches incorporate task-specific rewards, e.g., \citet{wu2018learning} learned models of coherence from existing text and used them as RL rewards; 
\citet{pasunuru2018multi} incorporated entailment-based consistency rewards to improve the saliency of a good summary;
\citet{ryu2024multi} optimized multiple dimensions such as relevance and fluency.
We extend this line of work by introducing a novel reference-free discourse-level reward that explicitly promotes topical alignment between summaries and source documents--a dimension overlooked by existing metrics. 
Unlike prior studies focused on single-document summarization, we address the more challenging multi-document setting, where certain proposed rewards (e.g., \citet{ryu2024multi}) become impractical.

%=================
\section{Incorporating Topic Labels into MDS via Prompting}
\label{sec:teaser}

Topic phrases succinctly capture essential information from source documents, providing effective high-level guidance for summarization--our motivation aligns with prior work on \textit{Entity Chains} \citep{narayan2021planning}, which utilized ordered sequences of entities as intermediate representations to plan and ground abstractive summary generation. 
However, unlike the controlled entity sets used in entity chains, we treat our topics as open-ended keywords and phrases.
Additionally, rather than incorporating entity generation directly into conditional summarization, we adopt a two-step framework where the topic extraction model is separate from the summarization model. 
This modular design enables independent analysis of topic extraction's impact on summarization quality (see \cref{subsec:results-quali}), and provides the flexibility to incorporate more advanced topic models in future experiments.

In this section, we conduct experiments in a zero-shot setting, carefully designing prompts and 
configurations to best leverage topic-augmented MDS.

\paragraph{Prompting with Topics}
Formally, given a set of source documents with corresponding topic labels, we prompt a LLM for summarization as follows:
\begin{equation}
P(S | doc^1, T_{doc^1}, \dots, doc^K, T_{doc^K}; \theta),
\vspace{-1ex}
\end{equation}
where $T_{doc^k}$ denotes topic labels for document $k$, and $\theta$ represents the LLM parameters. We append each set of topic labels immediately after its corresponding document, providing explicit topical guidance to assist the summarization model. %in identifying and extracting key information. 
This resembles the \textit{summary-level entity plans} introduced in \citet{narayan2021planning}, but extends naturally to multiple document-topic pairs, a format we found consistently more effective than an aggregated-topic version in pilot experiments. 
Detailed prompt examples are provided in Appendix~\ref{append-teaser}.

\paragraph{\textit{Teacher-Supervision} Mode}
We examine LLM capabilities by comparing their performance of varying scales (Qwen2.5-0.5B, 1.5B, and 7B).
Unsurprisingly, the largest summarization model (7B) achieves the highest baseline performance (average \textsc{Rouge} $19.9$), significantly surpassing smaller models (see ``Base'' in Figure~\ref{fig:prompt-hs}). 
We employ a teacher-supervision mode, where the larger 7B model explicitly provides topic guidance for the smaller models (0.5B and 1.5B). Under this setting, smaller LLMs clearly benefit from improved topical information provided by the teacher model. However, the 7B model itself, which inherently possesses strong topical modeling capabilities, experiences no %performance 
gains from self-generated topic labels.

\paragraph{Number of Topic Labels}
We also explore how the number of topic labels impacts summarization effectiveness, comparing summaries guided by $1$, $5$, or $10$ topics. 
A single topic label overly constrains summarization, leading to poorer performance across all models (``T1'' in Figure~\ref{fig:prompt-hs}). 
On the other hand, summarization quality notably improves when using %five or ten topic labels, 
more labels (``T5'' and ``T10''),
particularly for smaller models. 

These findings together suggest that employing richer topic signals through teacher-supervised extraction is beneficial, 
motivating us to incorporate topic information into the learning process.

%=================
\section{LLM Reinforcement Learning for MDS}
\label{sec:method}

Based on the above observations, we propose a novel topic-guided reward (\cref{subsec:topic-reward}) designed to maximize semantic similarity between generated summaries and source documents, coupled with a length penalty  (\cref{subsec:length-reward}) to better control the generation length. 
We implement these rewards using an inverse standard deviation weighting strategy (\cref{subsec:reward-weighting}) through the recent Group Relative Policy Optimization (GRPO) framework (\cref{subsec:grpo-training}).
See Figure~\ref{fig:pipeline} for the overview of our pipeline.

\subsection{Topic-Guided Reward}
\label{subsec:topic-reward}

A key contribution of our approach is a \textbf{Topic-F1 reward} metric that can effectively capture the semantic alignment between summaries and their respective source documents. We utilize a two-step embedding and matching procedure to quantify coverage and precision of topics.

For one data input $d=\{doc^1, doc^2, \ldots, doc^K\}$, we first apply the Qwen2.5-7B model 
to extract a set of topic labels $T_{doc} = \{t_1, t_2, \dots,t_n\}$ 
from each source document $doc^k$.
We design our experiments using different topic numbers: 
topic number $n=|T_{doc}|=10$ for Multi-News \citep{fabbri2019multi} and $n=|T_{doc}|=5$ topics for Multi-XScience \citep{lu2020multi}. 
In preliminary experiments, we observed that inference performance varied with the number of topics, 
and the best results were achieved when the number of topics matched the number of sentences in the gold summary, aligning with prior work \citep{narayan2021planning}.
Each topic label--may be a single word or a short phrase--is converted into a dense embedding using the SentenceTransformer model \texttt{all-mpnet-base-v2} \citep{reimers2019sentence}. 
We select this model due to its compact size and proven effectiveness in generating high-quality %sentence 
embeddings, adding minimal computational overhead in training.

Given a generated summary $S_g$, we similarly extract and embed its topics $T_\text{sum} = \{t_1, t_2, \dots,t_m\}$. We construct a similarity matrix $M$, whose entries $M_{ij}$ represent the cosine similarity between topic embeddings of each pair of topic phrases from source document and generated summary:
\vspace{-1ex}
\begin{equation}
M_{ij} = \frac{\mathbf{e}_\text{doc,i} \cdot \mathbf{e}_\text{sum,j}}{|\mathbf{e}_\text{doc,i}| |\mathbf{e}_\text{sum,j}|},
\end{equation}
where $\mathbf{e}_\text{doc,i}$ and $\mathbf{e}_\text{sum,j}$ represent embeddings for the $i^{th}$ document topic and $j^{th}$ summary topic, respectively. 
Note that the number of extracted topics from the source document and the generated summary may differ, as summaries are typically much shorter than source documents. 
We set the number of topics $m=|T_\text{sum}|=5$ for both datasets.

Then, we define \textbf{Coverage} as the average of the maximum similarity scores between each source topic and its most similar summary topic. Conversely, \textbf{Precision} is defined as the average of the maximum similarity scores between each summary topic and its most similar source topic:
\begin{equation}
\text{Coverage} = \frac{1}{n}\sum_{i=1}^{n}\max_{j = 1, 2, \ldots, m}(M_{ij}),
\end{equation}
\begin{equation}
\text{Precision} = \frac{1}{m}\sum_{j =1}^{m}\max_{i = 1, 2, \ldots, n}(M_{ij}).
\end{equation}

Finally, we calculate the \textbf{harmonic mean} of coverage and precision to derive our topic-guided reward $r_\text{topic}$.
This metric is computed pairwise for every source document-summary pair, encouraging generation of generic summaries that consistently capture key semantic elements across multiple documents. 
The final reward score $R_\text{topic}$ is obtained by averaging the $r_{topic}$ values across all document-summary pairs of one data point.
Preliminary experiments revealed that computing topic rewards on a pairwise basis consistently outperformed approaches that first merged topics across all documents before comparison, motivating our choice of topic alignment calculation.

\subsection{Length-Penalty Reward}
\label{subsec:length-reward}

As recent research shows, LLMs often fail to respect desired length constraints specified in prompts \citep{stiennon2020learning, wang2024positionid}. To mitigate excessive long (or short) output, we introduce a \textbf{token-level length reward} designed to penalize deviations from the target length. 
To determine the number of tokens, we use the tokenizer associated with the reference model--specifically, the Qwen2.5-0.5B model in our case.

Formally, the length reward $R_\text{len}$ is defined as:
\begin{equation}
R_\text{len} = \exp\left(-\frac{|L_\text{exp}-L_\text{sum}|}{L_\text{exp}}\right),
\end{equation}
where $L_\text{exp}$ represents the desired summary length and $L_\text{sum}$ the generated summary length. We compute $L_\text{exp}$ on a small validation set, with its size tunable to reflect user preferences.

In our pilot experiments, we evaluated both sentence-level and token-level approaches for length penalty. The results showed that token-level control led to significantly better adherence to the target length, effectively preventing summaries from becoming excessively long (up to five times the target length observed in initial trials).

\subsection{Reward Weighting}
\label{subsec:reward-weighting}

Our reward formulation can be viewed within the broader Multi-Objective Reinforcement Learning (MORL) framework, where multiple objectives--topic precision, coverage, and length constraints--must be simultaneously balanced. 
Inspired by the MORL literature \citep{roijers2013survey, van2017hybrid} and adaptive weighting strategies such as leveraging reward variance \citep{kendall2018multi}, 
we adopt an \textbf{inverse standard deviation weighting} scheme to stabilize training signals. 
Given reward signals $R_r$ with standard deviations $\sigma_r$ which we obtain from a mini-batch (approx. $5\%$ of training set), where $r$ refers to the reward type, the initial weights are defined as:
\vspace{-1.3ex}
\begin{equation}
w_r = 1 / \sigma_r.
\vspace{-1.5ex}
\end{equation}

Additionally, following %common practice 
\citet{stiennon2020learning}, we apply an emphasis factor of 2 to the topic-guided reward to reflect domain-specific priorities. This factor is a tunable hyperparameter, selected based on development set performance.
The final weights are normalized across all reward types:
\vspace{-1.3ex}
\begin{equation}
w_r^{\text{norm}} = \frac{w_r \times \text{factor}_r }{\sum_k (w_k \times \text{factor}_k)},
\vspace{-1ex}
\end{equation}
where $\text{factor}_\text{topic} = 2$ and $\text{factor}_\text{len} = 1$.
In further experiments,
we incorporate the reference-based \textsc{Rouge} reward alongside our reference-free topic-F1 reward, assigning equal weighting to both.
This strategy efficiently balances multiple reward components and dynamically emphasizes key metrics.

\subsection{GRPO Training}
\label{subsec:grpo-training}

To integrate our weighted reward into GRPO training \citep{shao2024deepseekmath}, we construct a scalar value $R_\text{total}$ which combines topic-F1 and length rewards:
\begin{equation}
R_{\text{total}}(S_g) = \sum_{r} w_r^{\text{norm}} R_r(S_g).
\end{equation}

The GRPO algorithm computes relative advantages of $R_\text{total}$ within a group of 
$G$ sampled 
completions, i.e., generated summaries:
\begin{equation}
A^{\text{GRPO}}_\text{g}=\frac{R_{\text{total}}(S_g)-\frac{1}{G}\sum_{g=1}^G R_{\text{total}}(S_g)}{\operatorname{std}_{g=1,2,\ldots,G}(R_{\text{total}}(S_g))}.
\end{equation}
Given this advantage estimation, the training objective is to optimize the policy ($\pi$) parameters $\theta$ by maximizing a clipped surrogate objective:

\vspace{-1.5ex}
\small
\begin{equation}
\begin{aligned}
L^{\text{GRPO}}(\theta) = \mathbb{E}_{S_g\sim\pi_{\theta_{\text{old}}}}\Bigl[ & \frac{1}{G}\sum_{g=1}^G
\min \Bigl(r_g(\theta) A_g, \\
&\text{clip}(r_g(\theta), 1-\epsilon, 1+\epsilon) A_g\Bigl) \Bigr] \\
& - \beta \cdot \mathbb{D}_{\mathrm{KL}}(\pi_{\theta}\|\pi_{\mathrm{ref}})
\end{aligned}
\end{equation}
\normalsize
with probability ratio \( \textstyle r_g(\theta)=\frac{\pi_{\theta}(S_\text{g}|d)}{\pi_{\theta_{\text{old}}}(S_\text{g}|d)} \) and a KL penalty to regularize policy updates.
Note however that our reward design is agnostic to specific RL algorithms, 
we adopt the GRPO framework due to its recent success 
\citep{shao2024deepseekmath, guo2025deepseek} and computational efficiency by removing the value model. 

%=================
\section{Experimental Setup}

\subsection{Datasets}

We choose two popular MDS datasets whose source documents and summaries different along multiple facets such as length and abstractiveness. 
% The key statistics of datasets are shown in Table~\ref{tab:datasets}.
\textbf{(1) Multi-News} \citep{fabbri2019multi} is one of the most widely used MDS datasets in news domain. It contains in average $2.7$ source documents per summary with relatively long documents. 
% Its reference summaries are approximately ten sentences long. 
\textbf{(2) Multi-XScience} \citep{lu2020multi} comprises the abstract of a query paper and those of its cited papers as input, with the goal of generating a related work paragraph. % This setup diverges from traditional MDS tasks. 
On average, it includes $4.4$ source documents and has highly abstractive summaries, making it particularly challenging. %for MDS models.
% \textcolor{blue}{
The key statistics of both datasets are given in Table~\ref{tab:datasets} of Appendix~\ref{append-datasets}.
% }

\subsection{Evaluation Metrics}
We report several complementary metrics that examine different aspects of the generated summaries. 
To assess summary \textit{informativeness},
we use lexical overlap metrics (e.g., \textsc{Rouge}-1/2/L/geometric mean; \citep{lin2004rouge}), along with embedding-based semantic similarity measures including \textsc{BERTScore} (short in \textsc{Bert}; \citealp{zhang2019bertscore}) and \textsc{Llm2vec Score} (short in \textsc{Llm2v}; \citealp{behnamghaderllm2vec}). The \textsc{Llm2v} metric we use is built upon the \texttt{Meta-LLaMA-3-8B} model fine-tuned \citep{llama3} with unsupervised contrastive learning \citep{gao2021simcse}.
Additionally, we examine \textit{topical alignment} via \textsc{CovRatio} and \textsc{PreRatio}, reflecting respectively the coverage and precision of extracted topics between the summary and source documents.

\subsection{Model Comparisons}
We primarily use Qwen series for our experiments \citep{qwen2025qwen25technicalreport}.
For all model variants, Qwen-2.5 0.5B-Instruct is used as the policy model in RL training. 
For reward calculation, we compare different sizes and types of reward model.
For all RL-training, we include the length penalty described in \cref{subsec:length-reward}. 
Implementation details are in the Appendix~\ref{append-exp}.
We compare the following variants:

    \textbf{(1) RL-Trained, Topic-reward:} Our proposed method, training a policy model (0.5B) with topic-F1 reward and GRPO. We include \textsc{RL\textsubscript{Topic-7B}} which leverages Qwen-2.5 7B-Instruct model as
    % \textit{topic teacher} 
    topic extractor,
    and explore a smaller %\textit{self-taught} 
    variant \textsc{RL\textsubscript{Topic-0.5B}} with 0.5B model for topic extraction. 
    % \shafiq{Make it more specific and clear.} --> lisa: rephrased
    
    % \item[2.] 
    \textbf{(2) RL-Trained, Human-feedback:} We compare against a reward model trained to predict human preference  
    % \citep{ouyang2022training}
    % and DPO \citep{ouyang2022training, rafailov2023dpo} 
    from OpenAssistant\footnote{\url{https://huggingface.co/OpenAssistant/reward-model-deberta-v3-large-v2.}} (\texttt{deberta-v3-large-v2}): \textsc{RL\textsubscript{human-feedback}}. 
    % This reward is also implemented into GRPO.
    
    % \item[3.] 
    \textbf{(3) Base:} 
    % \textcolor{blue}{
    We consider two sizes of base model in the Qwen2.5 family: 0.5B model and 7B model evaluated in a zero-shot setting (\textsc{Base (0.5B)}, \textsc{Base (7B)}). We also compare against 0.5B model guided by topic labels provided by the 7B model: \textsc{Base\textsubscript{Topic-7B}}, where it approximates the \textit{Entity Chains} \citep{narayan2021planning} within LLMs.
    
    % \item[4.] 
    \textbf{(4) Supervised Fine-Tuning (SFT):} We fine-tune Qwen2.5-0.5B-Instruct  model for summary generation with the SFT objective.
    
    % \item[5.] 
    \textbf{(5) RL-Trained, Reference-based:} Finally, we implement \textsc{Rouge}-reward using the geometric mean of \textsc{Rouge}-1/2/L within GRPO framework: \textsc{RL\textsubscript{Rouge}}, and benchmark with our model which uses a combination of topic and \textsc{Rouge} rewards: \textsc{RL\textsubscript{Topic-7B+Rouge}}.
% \end{itemize}

\section{Results and Analysis}
\label{subsec:results}

\subsection{Reference-free Results}

\begin{table*}[t]
    \centering
    % \normalsize
    % \small
    \resizebox{\linewidth}{!}{
    \begin{tabular}{lllc cccc cc cc}
    \toprule
    &&&& \multicolumn{4}{c}{Overlap-Based} & \multicolumn{2}{c}{Similarity-Based} & \multicolumn{2}{c}{Topic Alignment} \\
    \cmidrule(lr){5-8} \cmidrule(lr){9-10} \cmidrule(lr){11-12} 
    & Model & IM & RM & Rouge-1 & Rouge-2 & Rouge-L & Rouge-M & \textsc{Bert} & \textsc{Llm2v} & \textsc{CovRatio} & \textsc{PreRatio} \\
    \midrule
    % \rowcolor[gray]{0.9}\multicolumn{9}{l}{\textit{Multi-News}} \\
    \rowcolor[gray]{0.9}\multicolumn{12}{l}{\textit{Reference-free methods}} \\
    \multirow{6}{*}{\rotatebox[origin=c]{90}{News}} & 
    \textsc{Base (0.5B)} & 0.5B & - & $27.22$ & $7.28$ & $15.03$ & $14.31$ & $.842$ & $.721$ & $.513$ & $.622$ \\
    & \textsc{Base (7B)} & 7B & - & $37.09$ & $\underline{10.77}$ & $\textbf{19.77}$ & $\underline{19.91}$ & $\textbf{.845}$ & $\underline{.796}$ & $\underline{.538}$ & $\underline{.672}$ \\
    & \textsc{Base\textsubscript{topic-7B}} & 0.5B & 7B & $28.62$ & $8.60$ & $15.83$ & $15.73$ & $\underline{.844}$ & $.733$ & $.521$ & $.632$ \\
    & \textsc{RL\textsubscript{human-feedback}} & 0.5B & 0.3B & $33.07$ & $6.99$ & $17.29$ & $15.58$ & $.819$ & $.706$ & $.492$ & $.583$ \\
    & \textsc{RL\textsubscript{Topic-0.5B}} (ours) & 0.5B & 0.5B & $\underline{38.63}$ & $10.72$ & $18.81$ & $19.82$ & $\textbf{.845}$ & $.793$ & $.536$ & $\underline{.672}$ \\
    & \textsc{RL\textsubscript{Topic-7B}} (ours) & 0.5B & 7B & $\textbf{39.62}$ & $\textbf{10.97}$ & $\underline{18.97}$ & $\textbf{20.20}$ & $\textbf{.845}$ & $\textbf{.798}$ & $\textbf{.540}$ & $\textbf{.676}$ \\
    
    \hdashline
    \multirow{6}{*}{\rotatebox[origin=c]{90}{XScience}} & 
    \textsc{Base (0.5B)} & 0.5B & - & $25.05$ & $4.16$ & $13.47$ & $11.19$ & $.822$ & $.637$ & $.490$ & $.480$ \\
    & \textsc{Base (7B)} & 7B & - & $\underline{30.08}$ & $\underline{5.06}$ & $15.31$ & $\underline{13.26}$ & $\underline{.838}$ & $\underline{.728}$ & $\underline{.550}$ & $\underline{.549}$ \\
    & \textsc{Base\textsubscript{topic-7B}} & 0.5B & 7B & $25.62$ & $4.09$ & $13.93$ & $11.34$ & $.828$ & $.655$ & $.482$ & $.479$ \\
    & \textsc{RL\textsubscript{human-feedback}} & 0.5B & 0.3B & $26.78$ & $2.90$ & $13.87$ & $10.25$ & $.832$ & $.622$ & $.506$ & $.507$ \\
    & \textsc{RL\textsubscript{Topic-0.5B}} (ours) & 0.5B & 0.5B & $29.47$ & $4.79$ & $\underline{15.90}$ & $13.09$ & $.835$ & $.721$ & $.548$ & $\underline{.549}$ \\
    & \textsc{RL\textsubscript{Topic-7B}} (ours) & 0.5B & 7B & $\textbf{30.45}$ & $\textbf{5.38}$ & $\textbf{16.26}$ & $\textbf{13.86}$ & $\textbf{.847}$ & $\textbf{.741}$ & $\textbf{.554}$ & $\textbf{.560}$ \\ 
    \midrule
    
    \rowcolor[gray]{0.9}\multicolumn{12}{l}{\textit{Reference-based methods}} \\
    \multirow{3}{*}{\rotatebox[origin=c]{90}{News}} & \textsc{Sft} & 0.5B & - & $\underline{43.24}$ & $\underline{14.28}$ & $\underline{20.51}$ & $\underline{23.18}$ & $\underline{.852}$ & $\underline{.813}$ & $.529$ & $.665$ \\
    & \textsc{RL\textsubscript{Rouge}} & 0.5B & 0.5B & $41.43$ & $12.70$ & $19.19$ & $21.61$ & $.849$ & $.802$ & $\underline{.533}$ & $\underline{.670}$ \\
    & \textsc{RL\textsubscript{Topic-7B+Rouge}} (ours) & 0.5B & 7B & $\textbf{43.51}$ & $\textbf{14.31}$ & $\textbf{21.55}^*$ & $\textbf{23.40}$ & $\textbf{.857}^*$ & $\textbf{.823}^*$ & $\textbf{.543}^*$ & $\textbf{.683}^*$ \\
    
    \hdashline
    \multirow{3}{*}{\rotatebox[origin=c]{90}{XSci}} &
    \textsc{Sft} & 0.5B & - & $33.61$ & $\textbf{9.25}$ & $\textbf{18.28}$ & $\textbf{17.72}$ & $.850$ & $.750$ & $.480$ & $.510$ \\
    & \textsc{RL\textsubscript{Rouge}} & 0.5B & 0.5B & $35.20$ & $8.32$ & $18.07$ & $17.43$ & $.849$ & $.755$ & $.542$ & $.543$ \\
    & \textsc{RL\textsubscript{Topic-7B+Rouge}} (ours) & 0.5B & 7B & $\textbf{36.16}^*$ & $8.96$ & $18.15$ & $17.71$ & $\textbf{.852}^*$ & $\textbf{.765}^*$ & $\textbf{.557}^*$ & $\textbf{.569}^*$ \\
    \bottomrule
    \end{tabular}}
    \caption{
    % \textcolor{blue}{
    Results of reference-free (top) and reference-based (bottom) methods.
    Inference model (IM) sizes and reward / topic model (RM) sizes are given. Except for \textsc{Base (7B)}, all IM for summary generation is Qwen2.5-0.5B.
    % }
    We report %overlap-based 
    \textsc{Rouge} scores (1/2/L/geometric mean) \citep{lin2004rouge},
    \textsc{BertScore} \citep{zhang2019bertscore}, and \textsc{Llm2v} score \citep{behnamghaderllm2vec}, computed against gold summary. 
    We assess topic alignment via coverage (\textsc{CovRatio}) and precision (\textsc{PreRatio}).
    % \textcolor{blue}{
    Scores are averaged over six runs: for each setting, we train two independent models and each model is evaluated with three random seeds.
    In the reference-based comparison, $^*$ means our score is significantly better than \textsc{Sft} (Wilcoxon signed-rank test with $p<0.05$).
    Best score per experiment setting is in \textbf{bold} and second best \underline{underlined}.
    }
    % }
    \label{tab:result1}
    \vspace{-1ex}
\end{table*}

Since our topic-guided approach does not rely on references, we compare it first with other reference-free methods.
As shown in the top part in Table~\ref{tab:result1}, 
across both datasets, our method consistently outperforms all baselines in terms of summary informativeness.
Specifically, on Multi-News, \textsc{RL\textsubscript{Topic-7B}} achieves superior embedding-based similarity scores ($.845$ for \textsc{BERTscore} and $.798$ for \textsc{Llm} score) compared to \textsc{RL\textsubscript{human-feedback}} ($.819$ \textsc{BERTscore} and $.706$ \textsc{Llm} score). 
Even our smaller topic-guided variant, \textsc{RL\textsubscript{topic-0.5B}}, notably surpasses this baseline in all metrics, illustrating the robustness and scalability of our topic-guided reward framework. 
Interestingly, our RL-trained model, despite being much smaller, performed comparably with the much larger 7B model (\textsc{Base (7B)}). This is promising: it suggests that with effective topical guidance, a much smaller LLM can rival or exceed the performance of a significantly larger one.
% }
A similar trend is observed for Multi-XScience. 

In addition, %to informativeness metrics, 
we introduce a novel topic alignment assessment that directly evaluates the semantic alignment between generated summaries and source documents, independently of gold reference summaries. 
Our topic-guided models demonstrate significant improvements on both datasets, achieving increases of $2$-$7$ points in Coverage and $4$-$8$ points in Precision compared to baseline models. These consistent enhancements highlight the value of integrating direct topical guidance into the summarization process.

\subsection{Reference-based Results}

We evaluate our combined rewards (\textsc{RL\textsubscript{topic-7B+rouge}}) against reference-based approaches, specifically supervised fine-tuning (\textsc{Sft}) and RL with \textsc{Rouge} rewards. 

As shown in the second part in Table~\ref{tab:result1}, our model consistently surpasses these baselines in both similarity metrics and topic alignment scores.
On Multi-News, 
our RL model %trained with multiple rewards 
demonstrates clear superiority across all evaluated metrics.
In the more challenging Multi-XScience dataset--characterized by a larger number of source documents and highly abstractive summaries, as shown in Table~\ref{tab:datasets}--our RL model
notably excels in capturing semantic similarity and topic alignment.
This highlights RL's capacity to develop comprehensive summarization strategies beyond simple token imitation.
Additionally, we perform the Wilcoxon signed-rank test across all evaluation metrics, and our method shows statistically significant improvements over SFT in multiple metrics (scores marked with $^*$ in the Table).
On the other hand, \textsc{RL\textsubscript{rouge}} underperforms \textsc{Sft} in terms of \textsc{Rouge} scores, but performs on par with or even surpasses \textsc{Sft} in similarity-based and topic alignment metrics. This may be attributed to RL's enhanced exploration and generalization capabilities, which allow it not merely to mimic the next token, but to discover more effective generation patterns \citep{paulus2018deep}.

\subsection{RL Combined with Best-of-$n$ Strategy}
\begin{table*}[t]
    \centering
    % \small
    \resizebox{\linewidth}{!}{
    \begin{tabular}{ll cccc cc ccc}
    \toprule
    && \multicolumn{4}{c}{Overlap-Based} & \multicolumn{2}{c}{Similarity-Based} & \multicolumn{3}{c}{Topic Alignment} \\
    \cmidrule(lr){3-6} \cmidrule(lr){7-8} \cmidrule(lr){9-11} 
    & Model & Rouge-1 & Rouge-2 & Rouge-L & Rouge-M & \textsc{Bert} & \textsc{Llm2v} & \textsc{CovRatio} & \textsc{PreRatio} & \textsc{F1}\\
    \midrule
    \multirow{4}{*}{\rotatebox[origin=c]{90}{News}} &
    \textsc{Base (0.5B)} & $27.22$ & $7.28$ & $15.03$ & $14.31$ & $.842$ & $.721$ & $.513$ & $.622$ & $.562$ \\
    & \textsc{Base (0.5B)} + \textit{best-of-n} & $29.27$ & $8.68$ & $15.87$ & $15.92$ & $\textbf{.847}$ & $.738$ & $.517$ & $.647$ & $.575$ \\
    & \textsc{RL\textsubscript{Topic-7B+Rouge}} & $39.62$ & $10.97$ & $18.97$ & $20.20$ & $.845$ & $.798$ & $.540$ & $.676$ & $.600$ \\
    & \textsc{RL\textsubscript{Topic-7B+Rouge}} + \textit{best-of-n} & $\textbf{40.95}$	& $\textbf{12.03}$	& $\textbf{19.63}$	& $\textbf{21.30}$	& $.842$	& $\textbf{.798}$	& $\textbf{.546}$	& $\textbf{.683}$ & $\textbf{.607}$ \\
    
    \midrule
    \multirow{4}{*}{\rotatebox[origin=c]{90}{XScience}} & 
    \textsc{Base (0.5B)} & $25.05$ & $4.16$ & $13.47$ & $11.19$ & $.822$ & $.637$ & $.490$ & $.480$ & $.485$ \\
    & \textsc{Base (0.5B)} + \textit{best-of-n} & $27.88$	& $4.64$	& $14.68$	& $12.38$	& $.831$	& $.708$	& $.523$	& $.518$	& $.521$ \\
    & \textsc{RL\textsubscript{Topic-7B}} & $30.45$ & $5.38$ & $16.26$ & $13.86$ & $.847$ & $.741$ & $.554$ & $.560$ & $.557$ \\ 
    & \textsc{RL\textsubscript{Topic-7B}} + \textit{best-of-n} & $\textbf{30.94}$	& $\textbf{5.55}$	& $\textbf{16.37}$	& $\textbf{14.11}$	& $\textbf{.849}$	& $\textbf{.753}$	& $\textbf{.562}$	& $\textbf{.579}$	& $\textbf{.570}$ \\
    \bottomrule
    \end{tabular}}
    \caption{
    Combination of Best-of-$n$ Strategy with \textsc{Base} and our RL-trained models. On both datasets, $n=8$. We choose the response with the highest topic F1 score. Best score per dataset is in \textbf{bold}.
    }
    \label{tab:result3}
    % \vspace{-2ex}
\end{table*}

Test-time (or inference-time) scaling has emerged as a promising approach for enhancing LLM's performance without post-training, and has shown encouraging results for MDS \citep{cao2025multi2}.
Here, we investigate combining our RL method with the ``best-of-$n$'' strategy, which involves generating $n$ candidate outputs and selecting the one with the highest quality based on a chosen metric--here, our proposed topic F1 score. 

To evaluate the effectiveness of this strategy, we generate eight responses using different random seeds from the \textsc{Base} / RL-trained model, calculate all evaluation metrics, and select the response with the highest topic F1 score.
Results are given in Table~\ref{tab:result3}. On both datasets, the best-of-$n$ approach improves the Base model's performance. Notably, combining this strategy with our RL-trained model yields the best overall results. The performance trend is clear:
\textbf{$\text{RL + best-of-$n$} > \text{RL} > \text{Base + best-of-$n$} > \text{Base}$}.

These findings are encouraging, which suggest that: (1) Our proposed topic F1 metric is a reliable indicator for good summary in inference-time selection; (2) The best-of-n strategy and our RL-based training method are complementary and can be combined to achieve optimal performance.

\subsection{Results on Varying Source Documents}

\begin{figure}[t]
    \centering
    \includegraphics[width=.91\linewidth]{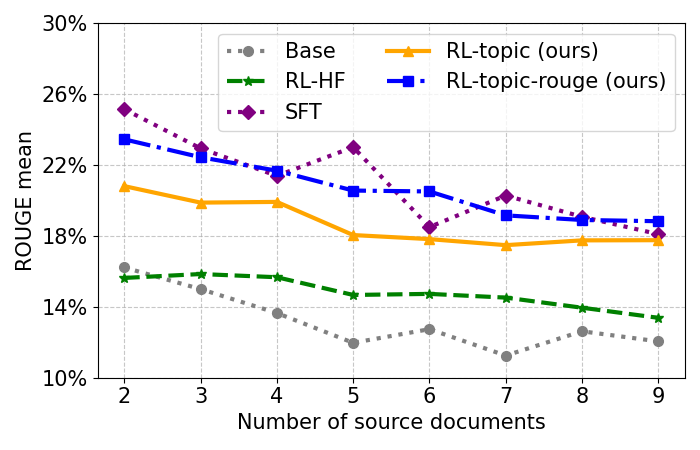}
    \caption{
    Model performance under different number of source document groups on Multi-News test set.
    }
    \label{fig:news-group}
\end{figure}

\begin{figure}[t]
    \centering
    \includegraphics[width=.91\linewidth]{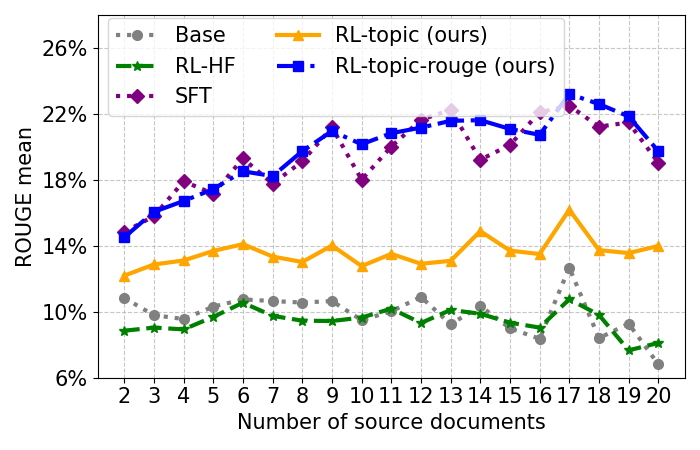}
    \caption{
    Model performance under different number of source document groups on Multi-XScience test set.
    }
    \label{fig:xsci-group}
    \vspace{-1ex}
\end{figure}

It is worth exploring how the number of source documents influences model performance. %To this end, we compare two datasets in terms of source 
We display the performance across different document number groups (with distribution given in Table~\ref{tab:stats-group}) of Multi-News and XScience datasets in Figures~\ref{fig:news-group} and \ref{fig:xsci-group}, respectively. 
We report the geometric mean of \textsc{Rouge} scores for comparison among \textsc{Base}, \textsc{RL\textsubscript{HF}}, \textsc{Sft}, and our two models: \textsc{RL\textsubscript{topic-7B}} and \textsc{RL\textsubscript{topic+rouge}}.

For Multi-News, \textsc{Rouge-M} scores decline as document number increases, confirming the challenge posed by multiple long documents. Although \textsc{Sft} achieves top performance on two-source documents, it exhibits significant instability and performance degradation with additional source documents. In contrast, our approaches exhibit more stable performance compared to all competitors. 

Multi-XScience reveals a contrasting trend: encouragingly, our models steadily improve performance with increasing numbers of source documents, a trend not observed with \textsc{Base} or \textsc{RL-HF}. Although \textsc{Sft} also shows improvement with more documents, its results fluctuate significantly, making it less reliable. Our RL-trained models, enhanced by topical information aligned with each source document, deliver the most consistent and superior performance, demonstrating clear advantages in practical multi-document scenarios.

\subsection{LLM-as-a-Judge Evaluation}

\begin{table}[]
    \centering
    \resizebox{\columnwidth}{!}{
    \begin{tabular}{lcc}
    \toprule
    A/B test & Vote & Winner \\
    \midrule
    (1a) \textsc{RL\textsubscript{Topic-7B}} vs. \textsc{Base} & $79$ vs. $21$	&  \textsc{RL\textsubscript{Topic-7B}}\\
    (1b) \textsc{Base} vs. \textsc{RL\textsubscript{Topic-7B}}	& $31$ vs. $69$	& \textsc{RL\textsubscript{Topic-7B}}\\
    (2a) \textsc{Sft} vs. \textsc{Base}	& $75$ vs. $25$	& \textsc{Sft} \\
    (2b) \textsc{Base} vs. \textsc{Sft}	& $37$ vs. $63$	& \textsc{Sft} \\
    (3a) \textsc{RL\textsubscript{Topic-7B}} vs. \textsc{Sft}	& $64$ vs. $36$	& \textsc{RL\textsubscript{Topic-7B}} \\
    (3b) \textsc{Sft} vs. \textsc{RL\textsubscript{Topic-7B}}	& $43$ vs. $57$	& \textsc{RL\textsubscript{Topic-7B}} \\
    (4a) \textsc{RL\textsubscript{Topic-7B+Rouge}} vs. \textsc{Sft}	& $78$ vs. $22$	& \textsc{RL\textsubscript{Topic-7B+Rouge}} \\
    (4b) \textsc{Sft} vs. \textsc{RL\textsubscript{Topic-7B+Rouge}}	& $29$ vs. $71$	& \textsc{RL\textsubscript{Topic-7B+Rouge}} \\
    \bottomrule
    \end{tabular}}
    \caption{LLM-as-a-judge evaluation on Multi-News. Results show consistent preference for our method.}
    \label{tab:result4}
    \vspace{-2ex}
\end{table}

\begin{table}[h]
    \centering
    \resizebox{\columnwidth}{!}{
    \begin{tabular}{lcc}
    \toprule
    A/B test & Vote & Winner \\
    \midrule
    (1a) \textsc{RL\textsubscript{Topic-7B}} vs. \textsc{Base} & $57$ vs. $43$	&  \textsc{RL\textsubscript{Topic-7B}}\\
    (1b) \textsc{Base} vs. \textsc{RL\textsubscript{Topic-7B}}	& $46$ vs. $54$	& \textsc{RL\textsubscript{Topic-7B}}\\
    (2a) \textsc{Sft} vs. \textsc{Base}	& $29$ vs. $71$	& \textsc{Base} \\
    (2b) \textsc{Base} vs. \textsc{Sft}	& $89$ vs. $11$	& \textsc{Base} \\
    (3a) \textsc{RL\textsubscript{Topic-7B}} vs. \textsc{Sft}	& $90$ vs. $10$	& \textsc{RL\textsubscript{Topic-7B}} \\
    (3b) \textsc{Sft} vs. \textsc{RL\textsubscript{Topic-7B}}	& $17$ vs. $83$	& \textsc{RL\textsubscript{Topic-7B}} \\
    (4a) \textsc{RL\textsubscript{Topic-7B+Rouge}} vs. \textsc{Sft}	& $86$ vs. $14$	& \textsc{RL\textsubscript{Topic-7B+Rouge}} \\
    (4b) \textsc{Sft} vs. \textsc{RL\textsubscript{Topic-7B+Rouge}}	& $17$ vs. $83$	& \textsc{RL\textsubscript{Topic-7B+Rouge}} \\
    \bottomrule
    \end{tabular}}
    \caption{LLM-as-a-judge evaluation on Multi-XScience. Results show consistent preference for our method.}
    \label{tab:result4-2}
    \vspace{-1.5ex}
\end{table}

Due to the heavy workload of the multi-document processing, it is challenging to conduct extensive manual evaluations. 
As a promising alternative \citep{chiang2023large, laban2024summary}, we leverage powerful LLMs--precisely, GPT-4.1 \citep{achiam2024gpt4}, the latest version available at the time of the experiment--as our automatic evaluator for large-scale evaluation. 

We specifically chose pairwise comparisons because they are shown to be easier to respond to, both in humans \citep{shah2018simple} and automatic evaluators \citep{wang2024direct}. 
Our include four systems: \textsc{Base}, \textsc{Sft}, \textsc{RL\textsubscript{Topic-7B}}, and \textsc{RL\textsubscript{Topic-7B+Rouge}}.
For each dataset, we conduct four pairwise comparisons:
(1)~\textsc{RL\textsubscript{Topic-7B}} vs. \textsc{Base};
(2)~\textsc{Sft} vs. \textsc{Base};
(3)~\textsc{RL\textsubscript{Topic-7B}} vs. \textsc{Sft};
(4)~\textsc{RL\textsubscript{Topic-7B+Rouge}} vs. \textsc{Sft}.

To reduce potential position bias, we conduct each pairwise comparison twice, switching the order of the two models in the A/B tests. Our judge prompt is adapted from \citep{xu2025j4r}, 
% instructs the LLM judge to ``evaluate which answer is more topically aligned with the source documents'' 
with details in Appendix~\ref{append-llm-judge}. 
We randomly sample $100$ examples per dataset, and the results are reported in Table~\ref{tab:result4} and \ref{tab:result4-2}. 
% Multi-XScience has similar observations, see supplementary material E.
We see that our %topic-guided 
RL-trained model consistently receives higher preference than both the \textsc{Base} and \textsc{Sft} models, providing strong evidence that our RL model performs better in terms of overall topic alignment.

\subsection{Qualitative Analysis}
\label{subsec:results-quali}

\paragraph{Human Evaluation on Topic Quality}
To verify our hypothesis that explicitly extracted topic phrases can effectively guide MDS, we conduct a human evaluation assessing the quality of topics generated by Qwen 7B and Qwen 0.5B models. 
Specifically, we evaluated four criteria—\textit{Relevance}, \textit{Coverage}, \textit{Specificity}, and \textit{Redundancy}—using a 5-point Likert scale. 
Detailed evaluation guidelines and results are provided in Appendix~\ref{append-human}.

In brief, evaluation results indicate that the 7B model consistently produces precise and conceptually rich topic phrases, often comprising multi-word expressions. 
In contrast, the 0.5B model tends to generate topic that, while relevant, lack sufficient coverage and specificity, and with semantic redundancy.
These findings support the benefits of \textit{teacher-supervised} framework, where larger models with superior topic-modeling capabilities effectively guide smaller models through topic distillation, thereby improving summarization performance.

\paragraph{Failure Cases in Generation}
During evaluation, we observe that %despite specifying a maximum output length during training, 
models occasionally produce excessively long and repetitive outputs. %at inference time. 
We quantify the frequency of such failure cases across all model variants (see Appendix~\ref{append-quali}) and find that the \textsc{Sft} model is most prone to this issue, with over $3\%$ of instances failing to generate coherent sentences. This partly accounts for the high variance observed in its performance. 
In contrast, the RLHF-trained model, as well as our proposed models with topic cues, exhibit greater stability, with minimal occurrence of such degenerate outputs ($<0.2\%$).

%=================
\section{Conclusion}

We introduce an interpretable, reference-free topic-guided RL approach for MDS, leveraging a novel topic-F1 reward that aligns summary topics with source documents. 
Integrated within the GRPO framework, our method consistently outperformed strong baselines, demonstrating the value of explicit topic guidance.
Looking forward,
we aim to enrich our framework by exploring advanced neural topic modeling techniques \citep{bianchi2021, fang2024cwtm} for more refined topical guidance. Moreover, incorporating innovative reward signals, such as LLM-as-a-judge evaluation \citep{zheng2023judging, liusie2024llm}, could further align summaries with human preferences and enhance self-consistency. Extending our topic-guided approach to interactive, query-based scenarios--where users specify key points to summarize--also presents an exciting future direction.

%=================
\section*{Limitations}

In our experiments, we focus primarily on models from the Qwen series, selected for their strong performance across diverse NLP tasks and the availability of multiple model sizes. 
This choice enables us to highlight and isolate the impact of topic alignment, avoiding potential confounding factors from different architectures across different LLM families. 
Furthermore, the policy model employed in our current setup is a 0.5B parameter model. Though exploring larger models is promising, the substantial computational cost limits such experiments in the current study. 
Nonetheless, our results clearly demonstrate the effectiveness of the topic reward approach even with this modestly sized model, laying a solid foundation for future studies that may scale to more powerful models.

Evaluating text summarization continues to be challenging due to the multifaceted aspects involved in assessing summary quality \citep{kryscinski2019neural, fabbri2021summeval, goyal2022news}. In our work, we employ a range of automatic metrics, including traditional methods (\textsc{Rouge}), embedding-based approaches, and our newly proposed topic alignment metrics, which notably do not require reference summaries.

Although we acknowledge the availability of other reference-free metrics, integrating them effectively into our task--summarization of multiple lengthy source documents--is nontrivial. For example, our preliminary analysis with an entailment-based metric \textsc{SummaC\textsubscript{conv}} \citep{laban2022summac} revealed that factual scores assigned to gold-standard summaries were sometimes lower than those assigned to zero-shot prompted summaries. 
Upon careful inspection, we discovered this occurred because certain prompted summaries heavily mirrored the first paragraph of source texts, resulting in disproportionately high entailment scores at the sentence level. 
This scoring pattern, however, does not accurately reflect comprehensive summary quality, as a good summary must synthesize information distributed across multiple documents. 
Thus, we propose our topic coverage and precision scores as a more balanced evaluation approach tailored for this task.

\section*{Ethical Statement}

We have taken proactive steps to address ethical concerns related to our research. 
Our corpora were carefully selected to minimize potential issues with biased or hateful content. 
For human evaluation, we clearly instructed annotators to remain vigilant and identify any biased or inappropriate language within the data. The annotators
% , two graduate students majoring in Computer Science, 
participated voluntarily without specific compensation; however, they were encouraged to use the results of their evaluation work for their academic studies. 

Reward hacking remains a boarder challenge in LLM post-training and fine-tuning, where models may exhibit undesirable behaviors such as overusing specific reward-triggering words without producing coherent or consistent summaries. 
In our experiments, two of the authors manually examined $200$ randomly sampled summaries generated by our RL-trained models. We did not find such a pattern where the model excessively repeats topic-related words. 
Additionally, our use of a length-penalty reward appears to mitigate such behavior to some extent, as it encourages the model to terminate generation with an END token within a reasonable length, thereby reducing the likelihood of repetitive outputs, such as looping on a single topic-related token. 
Exploring more targeted ``guard rewards'' to prevent such reward hacking represents a interesting direction for future work.

\section*{Acknowledgments}
The authors thank the anonymous reviewers and the Area Chair for their valuable feedback and suggestions.
The authors acknowledge the support of the Natural Sciences and Engineering Research Council of Canada (NSERC).
Nous remercions le Conseil de recherches en sciences naturelles et en génie du Canada (CRSNG) de son soutien.

\bibliography{custom}

%%%%%%%%%%%%%%%%%%%%%%
\appendix

\section{Prompt Template for MDS and Topic Modeling}
\label{append-teaser}

We present various prompts used in our work, both in zero-shot setting (Table~\ref{tab:prompt1}), and RL training (Table~\ref{tab:prompt2}, \ref{tab:prompt3}, and \ref{tab:prompt4}).

\begin{table}[h]
    \centering
    % \vspace{-0.8ex}
    \scalebox{1.0}{
        \begin{tcolorbox}[
        title=MDS with Topic Labels Prompt in Zero-shot
        ]
    % \vspace{-0.8ex}
    A conversation between User and Assistant. The user provides news articles and topic labels, and the Assistant produces a short summary. The summary contains no more than **ten sentences** and **only** based on information from the provided articles and topic labels.\\
    Documents: \{\texttt{Doc 1 text}\}, \{\texttt{Doc 1 topics}\}, \{\texttt{Doc 2 text}\}, \{\texttt{Doc 2 topics}\}, $\dots$, \{\texttt{Doc K text}\}, \{\texttt{Doc K topics}\}. 
    % Assistant: 
    % \vspace{-0.8ex}
    \end{tcolorbox}}
    \caption{Prompt template used by the Qwen2.5-0.5B / 1.5B / 7B in preliminary zero-shot experiments to test the performance with integrated topic labels (See \cref{sec:teaser}).}
    \label{tab:prompt1}
\end{table}

\begin{table}[h!]
    \centering
    % \vspace{-0.8ex}
    \scalebox{1.0}{
        \begin{tcolorbox}[
        % colback=blue!5!white,
        % colframe=blue!75!black,
        title=MDS Prompt in RL training (Multi-News)
        ]
    % \vspace{-0.8ex}
    A conversation between User and Assistant. The user provides news articles, and the Assistant produces a short summary. The summary contains no more than **ten sentences** and **only** based on information from the provided articles.\\
    Documents: \{\texttt{Doc 1 text}\}, \{\texttt{Doc 2 text}\}, $\dots$, \{\texttt{Doc K text}\}
    % Assistant: 
    % \vspace{-0.8ex}
    \end{tcolorbox}}
    \caption{Prompt template used by Qwen2.5-0.5B to generate summary (rollout) from Multi-News during RL training.}
    \label{tab:prompt2}
\end{table}

\begin{table}[h!]
    \centering
    % \vspace{-0.8ex}
    \scalebox{1.0}{
        \begin{tcolorbox}[
        % colback=blue!5!white,
        % colframe=blue!75!black,
        title=MDS Prompt in RL training (Multi-XScience)
        ]
    % \vspace{-0.8ex}
    The user provides scientific articles, and the Assistant generates a related work paragraph based on the query paper's abstract and the abstracts of its referenced papers. The answer includes citations for all referenced papers (@cite\_id) and be approximately **five sentences long**. \\
    Documents: \{\texttt{Doc 1 text}\}, \{\texttt{Doc 2 text}\}, $\dots$, \{\texttt{Doc K text}\}
    % Assistant: 
    % \vspace{-0.8ex}
    \end{tcolorbox}}
    \caption{Prompt template used by Qwen2.5-0.5B to generate summary (rollout) from Multi-XScience during RL training.}
    \label{tab:prompt3}
\end{table}

\begin{table}[h!]
    \centering
    % \vspace{-0.8ex}
    \scalebox{1.0}{
        \begin{tcolorbox}[
        title=Topic Modeling Prompt in RL training
        ]
    % \vspace{-0.8ex}
    A conversation between User and Assistant. The user provides a news article, and the Assistant produces **five** key words or phrases as **topic labels**. The answer should be in the form of a list, with each item separated by a comma. Do not give any explanation or additional information.\\
    Document: \{\texttt{Doc text}\}
    % Assistant: 
    % \vspace{-0.8ex}
    \end{tcolorbox}}
    \caption{Prompt template used by the Qwen2.5-7B model to extract topic labels from the generated summaries (during reward calculation). Note that we pre-extract and store topic labels from the source documents before training, thus avoiding redundant topic extraction computations during the training process.}
    \label{tab:prompt4}
\end{table}

\section{Datasets Statistics}
\label{append-datasets}

In Table~\ref{tab:datasets}, we show key statistics of MDS datasets Multi-News and Multi-XScience.

\begin{table}[t]
    \centering
    \small
    % \resizebox{\columnwidth}{!}{
    \begin{tabular}{lcc}
    \toprule
    & Multi-News & Multi-XSci \\
    \midrule
    Nb. train data & $44,972$ & $ 30,369$\\
    Nb. test data & $5,622$ & $5,093$ \\
    Nb. refs per summ & $2.8$ & $4.4$ \\
    Avg. words / sents in docs & $2,103$ / $28$ & $942$ / $33$\\
    Avg. words / sents in summ & $263$ / $10$ & $116$ / $5$\\
    \% novel unigrams & $17.8$ & $57.1$\\
    \% novel bigrams & $42.3$ & $81.8$\\
    \bottomrule
    \end{tabular}
    \caption{Key statistics of Multi-News and Multi-XScience. These numbers show variance in the size of source documents (references per summary, avg. words and sentences) and difference in gold summary properties (novel n-grams).}
    \label{tab:datasets}
\end{table}

We also present the numbers of source documents in the two datasets.
As shown in Table~\ref{tab:stats-group}: Multi-News primarily features two-source documents, while Multi-XScience has a more balanced distribution, with 2-, 3-, and 4-source inputs together making up $52\%$ of test set.

\begin{table}[]
    \centering
    \small
    % \resizebox{\columnwidth}{!}{
    \begin{tabular}{lc}
    \toprule
    GRPO Hyperparameters & Value \\
    \midrule
    Training epoch & $2$ \\
    Number of processes & $6$ \\
    Max prompt length & $8092$ \\
    Max completion length & $1024$ \\
    Gradient accumulation steps & $21$ \\
    Number of generations & $8$ \\
    Per device train batch size & $4$ \\
    Learning rate & $1e-6$ \\
    KL Coefficient & $0.04$ \\
    Epsilon & $0.2$ \\
    Warm-up ratio & $0.1$ \\
    Temperature & $0.7$ \\
    \bottomrule
    \end{tabular}
    \caption{Hyperparameters for GRPO training.}
    \label{tab:hyperparams}
    % \vspace{-1.5ex}
\end{table}

\begin{table*}[]
    \centering
    \resizebox{\linewidth}{!}{
    \begin{tabular}{l ccccccccccccccccccc}
    \toprule
    Datasets &2&3&4&5&6&7&8&9&10&11&12&13&14&15&16&17&18&19&20 \\
    \midrule
    News & $54.6$ & $27.8$ & $10.9$ & $3.9$ & $1.7$ & $0.7$ & $0.2$ & $0.2$ & - & - & - & - & - & - & - & - & - & - & - \\
    XScience & $23.9$ &$13.3$ &$15.5$ &$10.6$ &$6.6$ &$6.0$ &$4.3$ &$2.9$ &$4.1$ &$2.9$ &$2.9$ &$1.8$ &$1.4$ &$1.3$ &$0.7$ &$0.6$ &$0.6$ &$0.3$ &$0.3$ \\
    \bottomrule
    \end{tabular}}
    \caption{Distribution of the number of source documents in Multi-News (News) and Multi-XScience (Science) datasets. Numbers are given in the test subset. Similar trends are observed for the training subsets.}
    \label{tab:stats-group}
    % \vspace{-1.5ex}
\end{table*}

\begin{table*}[t]
    \centering
    % \small
    \resizebox{\linewidth}{!}{
    \begin{tabular}{ll cccc cc cc}
    \toprule
    && \multicolumn{4}{c}{Overlap-Based} & \multicolumn{2}{c}{Similarity-Based} & \multicolumn{2}{c}{Topic Alignment} \\
    \cmidrule(lr){3-6} \cmidrule(lr){7-8} \cmidrule(lr){9-10} 
    & Model & Rouge-1 & Rouge-2 & Rouge-L & Rouge-M & \textsc{Bert} & \textsc{Llm2v} & \textsc{CovRatio} & \textsc{PreRatio}\\
    \midrule
    \multirow{3}{*}{\rotatebox[origin=c]{90}{News}}
    & \textsc{RL\textsubscript{Topic-F1}} & $40.40\textsubscript{.70}$	& $11.60\textsubscript{.56}$	& $19.43\textsubscript{.40}$	& $20.88\textsubscript{.59}$	& $.845\textsubscript{.0}$	& $.797\textsubscript{.0}$	& $.542\textsubscript{.0}$	& $.681\textsubscript{.01}$ \\
    & \textsc{RL\textsubscript{Topic-Coverage}} & $40.68\textsubscript{.21}$	& $10.70\textsubscript{.12}$	& $19.22\textsubscript{.60}$	& $20.29\textsubscript{.35}$	& $.847\textsubscript{.0}$	& $.793\textsubscript{.0}$	& $.543\textsubscript{.0}$	& $.671\textsubscript{.0}$ \\
    & \textsc{RL\textsubscript{Topic-Precision}} & $41.04\textsubscript{.16}$	& $11.18\textsubscript{1.6}$	& $19.34\textsubscript{1.1}$	& $20.68\textsubscript{1.4}$	& $.845\textsubscript{.01}$	& $.796\textsubscript{.01}$	& $.534\textsubscript{.0}$	& $.682\textsubscript{.0}$ \\
    
    \midrule
    \multirow{3}{*}{\rotatebox[origin=c]{90}{XSci}} & 
    \textsc{RL\textsubscript{Topic-F1}} & $30.61\textsubscript{.28}$	& $5.42\textsubscript{.12}$	& $16.28\textsubscript{.08}$	& $13.92\textsubscript{.17}$	& $.844\textsubscript{.01}$	& $.744\textsubscript{.01}$	& $.558\textsubscript{.0}$ & $.567\textsubscript{.01}$ \\
    & \textsc{RL\textsubscript{Topic-Coverage}} & $30.49\textsubscript{.22}$	& $5.22\textsubscript{.06}$	& $16.04\textsubscript{.01}$	& $12.99\textsubscript{.13}$	& $.836\textsubscript{.0}$	& $.732\textsubscript{.0}$	& $.554\textsubscript{.0}$	& $.555\textsubscript{.0}$ \\
    & \textsc{RL\textsubscript{Topic-Precision}} & $30.41\textsubscript{.16}$	& $5.33\textsubscript{.19}$	& $16.07\textsubscript{.09}$	& $13.35\textsubscript{.06}$	& $.831\textsubscript{.0}$	& $.725\textsubscript{.01}$	& $.545\textsubscript{.01}$	& $.578\textsubscript{.0}$ \\
    \bottomrule
    \end{tabular}}
    \caption{
    Ablation results on Multi-News (News) and XScience (XSci) datasets, where we use the F1 score, only topic coverage, and only topic precision as reward in model post-training.
    }
    \label{tab:result-ablation}
    \vspace{-1ex}
\end{table*}

\section{Training Details}
\label{append-exp}

\paragraph{Implementation}
We adapted TRL library\footnote{\url{https://github.com/huggingface/trl}} for GRPO training. 
Most of our experiments were conducted on %8 x NVIDIA RTX 4090 24GB GPUs, 
8 x NVIDIA A100 40GB GPUs,
where one GPU was dedicated for rollout, one for topic generation, and the rest for training.
We set the number of generations to $8$, per-device train batch size to $4$, gradient accumulation steps to $21$, and KL coefficient to $0.04$.
For rollout and topic generation, we used vLLM\footnote{\url{https://github.com/vllm-project/vllm}} for accelerated inference.

\paragraph{Hyper-Parameters}

Table~\ref{tab:hyperparams} lists the hyper-parameters for GRPO training.
%(top) and SFT (bottom). 
Our training was based on TRL, adapted to our datasets and compute constraints. The best-performing checkpoint is selected based on validation set.

Due to the significant computational demands of our experiments, extensive hyperparameter optimization was impractical. Instead, we conducted pilot small-scale tests, as described in the Section~\ref{sec:method}, to inform our experimental setup.

For example, we observed that GRPO training is highly sensitive to \textbf{learning rate} adjustments. Although previous literature suggests using moderately higher temperatures to facilitate exploration, we found that temperatures of $1e-5$ or higher caused considerable fluctuations during training, leading to gradient explosions. Consequently, we maintained a low learning rate of $1e-6$.

Another important observation relates to the \textbf{number of completions} per input sample. We noticed that increasing number of generated samples per input improved performance, aligning with findings reported in Open-r1. However, due to the long input length, increasing the number of completions further required enlarging the training batch size, resulting in out-of-memory (OOM) errors. Therefore, we selected a sample size of $8$, balancing performance gains and computational constraints.

\paragraph{Time Consumption}
To compute the topic-F1 reward, we needed to generate topics and perform matrix operations for alignment. We made efforts to optimize this process--reducing the reward computation time to $<2$ seconds per deta point. With this optimization, training still took approximately $48$ hours on Multi-News and $24$ hours on XScience, using 8× A100-40G GPUs for each run.

In comparison, training with other reward models such as human-preference scores can be even more computationally demanding, as discussed in the next paragraph.

\paragraph{Implementation with \textsc{Rlhf} Reward}
The \textsc{RL\textsubscript{human-feedback}} reward model is designed to predict a preference score between generated answers given a specific question. 
During our implementation, we observed that including full source documents and generated summary as input significantly increased the computation time for calculating preference scores ($>20$ seconds per data point), rendering it impractical for RL training. 

To address this issue, we utilized key topic phrases as a concise proxy for the original documents. This approach substantially reduced computation time while effectively preserving relevant source content. This experience also highlights that using RLHF trained rewards directly with lengthy documents is computationally prohibitive, whereas our proposed method introduces minimal computational overhead for evaluating topic alignment.

\section{Ablation Study on Topic Reward}
\label{append-ablation}
We conducted two ablation studies where the RL model is trained using only Coverage or only Precision as the topic reward signal, in comparison with using Topic-F1 as reward. 
For each variant, we generate three summaries using different random seeds. 
We report the mean and standard deviation across all evaluation metrics, as shown in Table~\ref{tab:result-ablation}.
We conclude two key observations:

(1) The RL model trained with the harmonic mean of Coverage and Precision consistently outperforms the single-metric models in ROUGE and embedding-based evaluations. This improvement is especially pronounced on the XScience, while on Multi-News, the results are comparable.

(2) Models trained with only one reward signal (Coverage or Precision) tend to optimize more that metric during training–a trend we observe during training–, and achieve higher values for the corresponding topic metric at the cost of the other. This tradeoff is evident in the evaluation: higher Precision comes with lower Coverage, and vice versa. In contrast, using the harmonic mean as a reward leads to a more balanced optimization of both metrics, resulting in better overall topic alignment.
Arguably, this balance is particularly important for challenging datasets like XScience, which involve multiple source documents and require highly abstractive summaries (see Table 1 in the main paper). In such difficult settings, the usage of harmonic mean reward proves especially effective to guide topic-aligned summary.

\section{Details of LLM-as-a-Judge Evaluation}
\label{append-llm-judge}

\paragraph{Prompt Template} 
Our judge prompt, adapted from \citep{xu2025j4r} (Appendix E.1), 
instructs the LLM judge to ``evaluate which answer is more topically aligned with the source documents'', as shown below:.

\begin{table}[h]
    \centering
    % \vspace{-0.8ex}
    \scalebox{1.0}{
        \begin{tcolorbox}[
        title=Judge Prompt 
        ]
    % \vspace{-0.8ex}
    Please act as an impartial judge and evaluate the quality of the summary provided by two models to the given multi-document summarization task displayed below. You will be given answers from Model A and Model B. Your task is to evaluate which answer is more topically aligned with the source documents. \\
    Focus on the **overall topical relevance** of each response, and choose the one that better aligns with the main themes of the source documents.
    Avoid any biases, such as order of responses, length, or stylistic elements like formatting.\\
    You should strictly follow the output format: \\
    1. Model A is better: [[A>B]] \\
    2. Model B is better: [[B>A]]
    % Assistant: 
    % \vspace{-0.8ex}
    \end{tcolorbox}}
    \caption{Prompt template used by the Qwen2.5-0.5B / 1.5B / 7B in preliminary zero-shot experiments to test the performance with integrated topic labels.}
    \label{tab:prompt5}
\end{table}

\paragraph{Further Discussion}
To elaborate on our observations during the LLM-as-a-judge evaluation, we provide a more detailed discussion in this section.
From Table~\ref{tab:result4} and \ref{tab:result4-2} in the main text,
we see that our topic-guided RL-trained model consistently receives higher preference than both the \textsc{Base} and \textsc{Sft} models. This provides strong evidence that our RL model performs better in terms of overall topic alignment, and confirms the validity of our proposed \textsc{CovRatio} and \textsc{PreRatio} metrics.

Interestingly, on the Multi-XScience dataset, the LLM judge prefers the \textsc{Base} model over the \textsc{Sft} model in terms of topic alignment. 
We observed that the \textsc{Base} model tends to ``copy'' content—particularly by heavily mirroring the first paragraph of the source documents. In contrast, the \textsc{Sft} model behaves as a rigid ``pattern learner'', often repeating citation placeholders like @cite\_x, which is characteristic of this dataset. The LLM judge appears to favor the \textsc{Base} model's copying behavior, likely because it results in summaries that appear smoother and more coherent. However, when compared to our RL-trained model, the LLM judge consistently prefers our model to \textsc{Base} or \textsc{Sft}. This also aligns with the results of our automatic metrics in the paper (see Table~\ref{tab:result1}): the \textsc{Sft} model shows low \textsc{CovRatio} and \textsc{PreRatio} values similar to the \textsc{Base} model, and both are substantially lower than our RL-trained model.

\begin{table}[t]
    \centering
    \small
    \begin{tabular}{p{1.6cm} p{5cm}}
    \toprule
    Criterion & Guiding Question \\
    \midrule
    Relevance & Do the phrases reflect the central themes or key ideas of the document? \\
    Coverage  &	Do the phrases collectively represent diverse and important parts? \\
    Specificity	 & Are the phrases informative and precise, not vague or overly general? \\
    Redundancy & Are any phrases repeated or semantically overlapping?\\
    \bottomrule
    \end{tabular}
    \caption{Evaluation criteria instructions.}
    \label{tab:instruct}
\end{table}

\begin{table}[t]
    \centering
    \small
    \resizebox{\columnwidth}{!}{
    \begin{tabular}{lcccc}
    \toprule
    Model & Relevance & Coverage & Specificity & Redundancy \\
    \midrule
    7B & $5.0$ & $5.0$ & $5.0$ & $5.0$ \\
    0.5B & $3.8$ & $2.8$ & $2.4$ & $3.2$ \\
    \bottomrule
    \end{tabular}}
    \caption{Model topic evaluation summary.}
    \label{tab:topic-eval}
    % \vspace{-2ex}
\end{table}

\section{Human Evaluation on Topic Quality}
\label{append-human}

We conducted a human evaluation to assess the quality of topic phrases generated by the Qwen2.5-7B and Qwen2.5-0.5B models. 
Two graduate-level annotators independently evaluated the outputs for ten randomly selected documents from the Multi-XScience training set. The evaluation was guided by four criteria--Relevance, Coverage, Specificity, and Redundancy--rated on a 5-point Likert scale (1 = poor, 5 = excellent). 
The detailed annotation instructions are provided in Table~\ref{tab:instruct}.

Annotators were first asked to read the source documents and were free to highlight or mark key phrases they considered important. 
Subsequently, they were presented with anonymized and randomly ordered topic lists generated by the two models. For each list, annotators assigned scores based on the four evaluation criteria.

The aggregated evaluation results for twenty topic sets (ten documents, two models) are summarized in Table~\ref{tab:topic-eval}. 
The findings indicate that Qwen2.5-7B consistently outperforms Qwen2.5-0.5B, producing more precise, accurate, and comprehensive topic phrases. In contrast, the 0.5B model exhibits notable deficiencies in coverage and specificity. 
Table~\ref{tab:topic-eval2} presents several representative examples to illustrate the qualitative differences between the models. 
In these examples, topics highlighted in red were identified by annotators as inappropriate, 
typically due to being overly generic or lacking clarity.

\begin{table}[]
    \centering
    \small
    \begin{tabular}{p{3.5cm} p{3.3cm}}
    \toprule
    Qwen-7B topics & Qwen-0.5B topics \\
    \midrule
    Communication strategies, collaborative problem solving, resource limitations, task requirements, experimental simulations & \textcolor{red}{Effective}, problem solving, resource bounded, communication, \textcolor{red}{collaborative} \\
    \midrule
    Principle of Parsimony, Task-Oriented Dialogue, Recovery Strategies, Information Transfer, HCRC Map Task & Parsimonious, task-oriented, \textcolor{red}{information}, recovery, dialogue \\
    \midrule
    Automatic Text Categorization, WordNet, Vector Space Model, Rocchio Algorithm, Widrow-Hoff Algorithm & WordNet, Rocchio, Widrow-Hoff, \textcolor{red}{category}, \textcolor{red}{low frequency} \\
    \midrule
    Natural Language Processing, TextTiling, TileBars, Cougar, Topic Labeling & Contextual, \textcolor{red}{text}, \textcolor{red}{topic}, retrieval, \textcolor{red}{display} \\
    \midrule
    Text categorization, WordNet, lexical databases, training collections, performance comparison & Auto text categorization, lexical databases, training collections, WordNet, \textcolor{red}{WordNet-based} \\
    \bottomrule
    \end{tabular}
    \caption{Examples of generated topics from Qwen2.5-7B and Qwen2.5-0.5B models. Topics highlighted in red are considered as inappropriate topics.}
    \label{tab:topic-eval2}
\end{table}

\section{Qualitative Results}
\label{append-quali}

During evaluation, we observe that despite specifying a maximum output length during training, 
models occasionally produce excessively long and repetitive outputs at inference time. 

We quantify the frequency of such failure cases across all model variants (see Table~\ref{tab:fail-percentage}) and find that the \textsc{Sft} model is most prone to this issue, with over $3\%$ of instances failing to generate coherent sentences.
This partly accounts for the high variance observed in its performance. 
In contrast, the RL-trained model with human preference rewards, as well as our proposed models with topic cues, exhibit greater stability, with minimal occurrence of such degenerate outputs ($<0.2\%$).

\begin{table}[]
    \centering
    % \small
    \resizebox{\linewidth}{!}{
    \begin{tabular}{lcc}
    \toprule
    Model & Multi-News (\%) & Multi-XScience (\%) \\
    \midrule
    \textsc{Base} & $1.37$ & $0.04$ \\
    \textsc{Base\textsubscript{topic}} & $2.41$ & $0.14$ \\
    \textsc{Sft} & $3.92$ & $3.14$ \\
    \textsc{RL\textsubscript{HF}} & $0.07$ & $0.00$ \\
    \textsc{RL\textsubscript{Rouge}} & $0.52$ & $0.02$ \\
    \textsc{RL\textsubscript{Topic-0.5B}} (ours) & $0.18$ & $0.00$ \\
    \textsc{RL\textsubscript{Topic-7B}} (ours) & $0.12$ & $0.00$ \\
    \textsc{RL\textsubscript{Topic+Rouge}} (ours) & $0.14$ & $0.08$ \\
    \bottomrule
    \end{tabular}}
    \caption{Percentage of failure cases %in test, 
    where model generates repetitive and long output (e.g., $>2,500$ tokens).}
    \label{tab:fail-percentage}
\end{table}

Below, we provide an example of two failure cases generated by the \textsc{Sft} model, corresponding to test example \#21 in the Multi-XScience and \#69 in Multi-XScience, respectively:

``\textit{
In recent years, many new methods have been developed to solve the blind image denoising problem. First, the mixture of Gaussian distribution @cite\_21 @cite\_8 @cite\_13 @cite\_30 @cite\_40 @cite\_10 @cite\_19 @cite\_23 @cite\_9 @cite\_32 @cite\_6 @cite\_25 @cite\_18 @cite\_23 @cite\_32 @cite\_18 @cite\_18 @cite\_13 ...}''
-- the model continues repeating content until it exhausts the maximum output length defined by vLLM.

``~\textit{
Russell Crowe has had enough of Mark Lawson's questions about his accent. Crowe, who's been accused of giving Robin Hood an Irish accent, walked out of a Radio 4 interview Wednesday, reports the Telegraph. I'm a little dumbfounded that you could possibly find any Irish in that character that's kind of ridiculous, but it's your show, he says. Crowe has been accused of making the movie's Robin Hood sound a little Irish, and Lawson tried to make a point about the accent. You've got dead ears, mate, he said. You've seriously got dead ears, if you think that's an Irish accent. Crowe, who lives in Australia, didn't seem to care. I'm a little dumbfounded that... [repeating]~}''.

\end{document}